\pdfoutput=1

\documentclass[11pt]{article}

\usepackage{acl}

\usepackage{times}
\usepackage{latexsym}
\usepackage{amssymb}
\usepackage{amsmath}
\usepackage{booktabs}
\usepackage{graphicx}
\usepackage{subcaption}
\usepackage{xcolor}

\usepackage{cleveref}
\crefformat{section}{\S#2#1#3}
\crefformat{subsection}{\S#2#1#3}

\usepackage[T1]{fontenc}

\usepackage[utf8]{inputenc}

\usepackage{microtype}

\title{Information Type Classification with Contrastive Task-Specialized Sentence Encoders}

\author{Philipp Seeberger, Tobias Bocklet, Korbinian Riedhammer \\
  Technische Hochschule Nürnberg \\
  \texttt{\{firstname.lastname\}@th-nuernberg.de}}

\begin{document}
\maketitle
\begin{abstract}

User-generated information content has become an important information source in crisis situations. 
However, classification models suffer from noise and event-related biases which still poses a challenging task and requires sophisticated task-adaptation. 
To address these challenges, we propose the use of contrastive task-specialized sentence encoders for downstream classification.
We apply the task-specialization on the \textsc{CrisisLex}, \textsc{HumAID}, and \textsc{TrecIS} information type classification tasks and show performance gains w.r.t. $F_{1}$ score.
Furthermore, we analyse the cross-corpus and cross-lingual capabilities for two German event relevancy classification datasets.

\end{abstract}

\section{Introduction}
\label{sec:introduction}

User-generated information content on social media has become an important information source in crisis and emergency situations \citep{reuter_2018}.
Social media posts immediately provide details about ongoing developments, first-party observations, and other information which would be missed with traditional sources (e.g., official news) \citep{sakaki_2010}.
Access to this information content is thereby crucial for situational awareness in order to support official institutions, government organisations, and relief providers \citep{kruspe_2021}.

However, processing this noisy high-volume social media streams is challenging and requires sophisticated methods for automatic reliable detection of information content.
To tackle this challenge, recent work has focused on binary, multi-class, and multi-label information type classification approaches \citep{crisismmd_2018,alam_2021,trecis2021}.
For instance, important information categories cover missing and injured people, damaged infrastructure, etc.

Another challenge is the nature of data prevalent in social media and microblogging platforms.
For example, a large portion of noisy user-generated texts inherit properties such as a limited number of words, less contextual information, hashtags, and noise (e.g., misspellings, emojis) \citep{wiegmann_2020,zahera_2021}.
Furthermore, event-related biases and entities prevent models from generalizing to unseen disaster events and therefore degrade in performance \citep{zhang2021,seeberger2022}.

These challenges motivates the use of efficient and effective approaches for adapting classifiers to the noisy text domain and the different information type tasks.
Recently, contrastive fine-tuning mechanisms attracted research efforts for few-shot and task-specialization settings by adapting language models and sentence encoders (SE) for downstream classification \cite{vulic2021,tunstall2022,su_2022}.
Following this approach, we aim to analyse the contrastive task-specialization in the field of information type classification.

\paragraph{Contributions}

Our main contributions are as follows: \textbf{1)} We introduce the contrastive task-specialization method for information type classification. \textbf{2)} We analyse the cross-corpus capabilities of the task-specialized models. \textbf{3)} We empirically show the cross-lingual and cross-task transfer capabilities for two German disaster datasets.

\section{Method}
\label{sec:method}

As discussed in section \ref{sec:introduction}, we follow previous work \cite{vulic2021} and aim to fulfill the requirement of effective adaptation by \textbf{1)} quickly bootstrapping a general-purpose SE for new domains and tasks via contrastive learning, and \textbf{2)} training a classifier on top of the fixed SE.

\subsection{Contrastive Task-Specialization}

The main idea is to follow the specialization of a general-purpose SE which is pre-trained on a large corpus of sentence pairs.
Specializing a universal SE to particular tasks has been proven effective in prior work for multi-class and multi-label scenarios via further fine-tuning by a contrastive loss \cite{vulic2021,zhang_contrast_2021,vulic2022}.
In this way, we can utilize available annotations to achieve task-adaptation to create more accurate encodings for the downstream classification.

\paragraph{Positive and Negative Pairs}

For the creation of sentence pairs, we follow prior efficient contrastive and few-shot approaches by implicitly leveraging the information type ids to create positive and negative learning examples \cite{tunstall2022}.
Therefore, we use the sentence pair creation scheme proposed by \textsc{SetFit}\footnote{\citet{tunstall2022} introduced \textsc{SetFit} as an efficient and prompt-free framework for few-shot classification and fine-tunes sentence transformers in a contrastive manner.} and construct the positive set $Pos$ and negative set $Neg$ by applying $n$ iterations of sentence pair generations.
For the multi-label task, we follow \cite{vulic2022} by sampling a positive sentence for each label in the label set of sentence $s_i$.

\paragraph{Contrastive Loss}

As contrastive loss, we opt for the Online Contrastive Loss (\textsc{OCL}) \cite{reimers2019,vulic2022}. 
This online version of contrastive loss operates with hard in-batch negative pairs and hard in-batch positive pairs and yields the final task-specialized SE.
The constrastive learning should attract similar sentences together and push dissimilar sentences apart.

\subsection{MLP Classification}

A standard approach for classification based on SE's is the Multi Layer Perceptron (MLP) which is stacked on top of a fixed SE.
This is much more lightweight than fine-tuning the entire SE but still achieves comparable performance in low-resource settings.
We train a MLP classifier composed of a single hidden layer with non-linearity.
For the multi-class and multi-label classifier, we use the standard cross-entropy and binary cross-entropy loss, respectively.
A threshold $\theta$ determines the final classification for the multi-label task by only classifying information types with probability scores $\geq \theta$.

\section{Experimental Setup}
\label{sec:experimentalsetup}

\subsection{Datasets}

We experiment with three \textsc{Twitter} datasets, covering \textbf{1)} multi-class and multi-label classification, \textbf{2)} different information type ontologies, and \textbf{3)} numerous diverse event types composed of natural disasters and human-made disasters.

\paragraph{\textsc{CrisisLex}}
The \textsc{T26} variant of \textsc{CrisisLex} \cite{olteanu2015} includes labeled tweets for 26 crisis events, annotated with seven information types including the category \textsc{Not Related}.
This set reflects a wide variety of events about emergencies with approximately 1,000 tweets per individual event.
As preprocessing step, we removed tweets with the label \textsc{Not applicable} as these contain issues such as "not readable" for the annotator \cite{olteanu2015}.
This task represents a multi-class classification problem.

\paragraph{\textsc{HumAID}}
This collection contains data about 19 events with dataset sizes ranging from approximately 570 to 9500 tweets \cite{Alam2021}.
\textsc{HumAID} covers eleven categories ranging from \textsc{Not Humanitarian} to \textsc{Inured or Dead People} which captures fine-grained information about disasters.
We ignore posts with the labels \textsc{Can't Judge} and \textsc{Missing or Found People} as the latter case is only available for four events.
Similar to \textsc{CrisisLex}, this task is about multi-class classification.

\paragraph{\textsc{TrecIS}}
TREC Incident Streams is a multi-label classification task composed of over 70 events with annotations for 25 information types \cite{trecis2021}.
The sample ranges are from 90 to 5900 tweets and highly vary across events in terms of tweets and label distribution.
Furthermore, the collection also covers the large-scale public-health event \textsc{COVID} whereby we only focus on general crisis events.
For our experiments, we drop \textsc{COVID} events and select the top-30 events with the highest number of posts as events with a few posts only cover a small subset of relevant information types.

\paragraph{Data Splits}

For within-corpus classification, we evaluate each method with 5-fold cross-validation (5-fold CV) with disjoint events.
Due to the high cost of task-specific annotations, we additionally focus on low-data scenarios for bootstrapping SE's.
Therefore, we conduct experiments in the two data configurations \textbf{1)} \textit{Low} and \textbf{2)} \textit{High}. For the \textit{High}-setup, we use the training and test splits provided by the 5-fold CV method. Then, in the \textit{low}-setup, we randomly sample 10 posts for each information type and event in order to construct the low-resource training sets.
Throughout all experiments, the test splits remain the same.

\begin{table*}[t]
    \centering
    \resizebox{\linewidth}{!}{
    \begin{tabular}{lcccccc}
        \toprule
        \multicolumn{1}{c}{} & \multicolumn{2}{c}{\textbf{\textsc{CrisisLex}}} & \multicolumn{2}{c}{\textbf{\textsc{HumAID}}} & \multicolumn{2}{c}{\textbf{\textsc{TrecIS}}} \\
        \cmidrule(rl){2-3} \cmidrule(rl){4-5} \cmidrule(rl){6-7}
         \textbf{Variant} & \textbf{Low} & \textbf{High} & \textbf{Low} & \textbf{High} & \textbf{Low} & \textbf{High} \\
        \toprule
        \textsc{MPNET\textsubscript{LM}+FFT} & \underline{54.5\hspace{0.3mm}$_{5.0}$} / \underline{46.7\hspace{0.3mm}$_{5.7}$} & \underline{62.9\hspace{0.3mm}$_{4.9}$} / \textbf{55.0\hspace{0.3mm}$_{3.6}$} & \underline{63.6\hspace{0.3mm}$_{4.0}$} / \underline{57.6\hspace{0.3mm}$_{3.6}$} & \textbf{73.2\hspace{0.3mm}$_{2.8}$} / \underline{66.3\hspace{0.3mm}$_{3.6}$} & 18.1\hspace{0.3mm}$_{1.2}$ / 14.3\hspace{0.3mm}$_{0.5}$ & 29.1\hspace{0.3mm}$_{2.1}$ / 22.8\hspace{0.3mm}$_{2.6}$ \\
        \cmidrule(rl){2-3} \cmidrule(rl){4-5} \cmidrule(rl){6-7}
        \textsc{MPNET\textsubscript{LM}} & 49.9\hspace{0.3mm}$_{4.7}^{*}$ / 42.9\hspace{0.3mm}$_{5.4}^{*}$ & 56.7\hspace{0.3mm}$_{5.9}^{*}$ / 48.6\hspace{0.3mm}$_{5.4}^{*}$ & 56.4\hspace{0.3mm}$_{4.3}^{*}$ / 52.2\hspace{0.3mm}$_{3.3}^{*}$ & 64.2\hspace{0.3mm}$_{5.8}^{*}$ / 59.5\hspace{0.3mm}$_{4.5}^{*}$ & \textbf{32.9\hspace{0.3mm}$_{3.0}^{*}$} / \underline{25.2\hspace{0.3mm}$_{3.1}^{*}$} & \textbf{30.4\hspace{0.3mm}$_{2.4}^{*}$} / \textbf{23.1\hspace{0.3mm}$_{3.3}$} \\
        \textsc{MPNET\textsubscript{SE}} & 51.7\hspace{0.3mm}$_{4.0}^{*}$ / 44.1\hspace{0.3mm}$_{4.8}^{*}$ & 56.7\hspace{0.3mm}$_{5.6}^{*}$ / 49.0\hspace{0.3mm}$_{5.8}^{*}$ & 58.4\hspace{0.3mm}$_{4.4}^{*}$ / 53.1\hspace{0.3mm}$_{2.6}^{*}$ & 65.6\hspace{0.3mm}$_{4.1}^{*}$ / 60.1\hspace{0.3mm}$_{3.2}^{*}$ & 32.1\hspace{0.3mm}$_{2.2}^{*}$ / 24.3\hspace{0.3mm}$_{3.7}^{*}$ & \underline{30.1\hspace{0.3mm}$_{2.3}$} / \underline{22.9\hspace{0.3mm}$_{3.6}$} \\
        \textsc{MPNET\textsubscript{SE}+CTS} & \textbf{56.6\hspace{0.3mm}$_{4.9}^{*}$} / \textbf{49.2\hspace{0.3mm}$_{5.5}^{*}$} & \textbf{63.0\hspace{0.3mm}$_{5.4}$} / \underline{54.3\hspace{0.3mm}$_{5.7}$} & \textbf{66.7\hspace{0.3mm}$_{3.1}^{*}$} / \textbf{60.6\hspace{0.3mm}$_{3.0}^{*}$} & \underline{72.7\hspace{0.3mm}$_{2.9}^{*}$} / \textbf{67.4\hspace{0.3mm}$_{3.7}$} & \underline{32.7\hspace{0.3mm}$_{3.1}^{*}$} / \textbf{25.3\hspace{0.3mm}$_{2.8}^{*}$} & 29.4\hspace{0.3mm}$_{3.0}$ / 22.6\hspace{0.3mm}$_{3.1}$ \\
        \bottomrule
    \end{tabular}}
    \caption{Overall results for event micro-averaged $F_{1}$ (x100\%) and macro-averaged $F_{1}$ (x100\%) scores with standard deviations. \textbf{Bold} numbers indicate the best performance whereas \underline{underlined} numbers denote the second best performance in each column. Results with $^{*}$ are significantly different from \textsc{MPNET\textsubscript{LM}+FFT} ($p$-value $ < 0.05$).}
    \label{table:results:overall}
\end{table*}

\subsection{Models and Hyperparameters}

For our evaluation, we use \textsc{MPNET\textsubscript{LM}}\footnote{\url{microsoft/mpnet-base}} \cite{song2020} as language model and \textsc{MPNET\textsubscript{SE}}\footnote{\url{sentence-transformers/all-mpnet-base-v2}} as SE variant, transformed by a standard contrastive dual-encoder framework.
\textsc{MPNET\textsubscript{LM}} comprises 12 transformer layers with hidden size $h_{T}=768$ and prior work has trained \textsc{MPNET\textsubscript{SE}} with approximately one billion sentence pairs.

\paragraph{Baseline}
As baseline, we additionally conduct full end-to-end fine-tuning of the \textsc{MPNET\textsubscript{LM}} model with a MLP classification head which we denote as \textsc{MPNET\textsubscript{LM}+FFT}.
Following suggested settings \cite{wang_2021,alam_2021,seeberger2022}, we train the baseline models for $15$ epochs with the optimizer AdamW, learning rate $2e-5$, weight decay $0.01$, batch size $32$, and evaluate the best checkpoint selected by a validation set.

\paragraph{Contrastive Task-Specialization}

In terms of CTS fine-tuning with OCL, we adopt a similar setup. 
The learning rate of AdamW is set to $2e-5$, weight decay to $0.01$, and batch size to $64$.
For the \textit{High}- and \textit{Low}-setup, we construct sentence pairs with $n=1$ and $n=5$, respectively.
We fine-tune the models on the sentence pairs for 3 epochs with the warmup ratio of 0.05 and cosine decay.

\paragraph{Classification}

The classifier consists of a MLP architecture with one hidden layer of size 512 with ReLU as non-linear activation function.
We train the classifier for 30 epochs with the opimizer AdamW, learning rate $1e-3$, weight decay $0.01$, dropout $0.4$, and batch size $32$.
For multi-label classification, we use the threshold $\theta$ of $0.3$.
We select the best classifier based on a validation split sampled from the training set with the ratio of $0.1$.

\subsection{Evaluation}

For all experiments, we report the micro-averaged and macro-averaged $F_{1}$ scores across events. 
In the \textit{High}-setup, all reported results are averaged across the five folds.
For the \textit{Low}-setup, we additionally conduct three runs with random seeds in order to reach more stable results with respect to few-shot sampling.

\section{Results and Discussion}

The main results are summarized in \tablename~\ref{table:results:overall}, while further cross-corpus and cross-lingual experiments are shown in \tablename~\ref{table:results:crosscorpus} and \tablename~\ref{table:results:crosslingual}. 
In the following, we discuss the results and findings.

\paragraph{Contrastive Task-Specialization}

The results in \tablename~\ref{table:results:overall} reveal that performance gains for the multiclass-classification are achieved via CTS.
These performance boosts are across all \textit{Low}- and \textit{High}-setups with respect to the \textsc{CrisisLex} and \textsc{HumAID} datasets.
With focus on the low-resource setup, we additionally experience significant improvements over the full fine-tuning baseline.
In comparison, the gap between \textsc{MPNET\textsubscript{SE}} and \textsc{MPNET\textsubscript{SE}+CTS} is consistently higher than the counterpart \textsc{MPNET\textsubscript{LM}} and \textsc{MPNET\textsubscript{SE}} which suggests the effectiveness of CTS.
However, there are no substantial performance gains or even a decrease for the \textsc{TrecIS} multi-label task.
This finding is contrary to the results in the domain of multi-label intent detection \cite{vulic2022}.
We hypothesize the cause are differences in semantic concepts across events and annotations \cite{seeberger2022}.
More sophisticated sentence pair sampling techniques, hard-negative mining or the usage of high level information types may tackle these shortcomings.

\paragraph{Cross-Corpus}

With cross-corpus evaluation we aim to analyze other important aspects of CTS fine-tuning.
We hypothesize that similar information type ontologies lead to better classification performances by transfering the fine-tuned knowledge about semantically similar information types.
Therefore, we trained the SE's on the source corpus and only trained the MLP classifier with the fixed SE on the target corpus.
The results in \tablename~\ref{table:results:crosscorpus} indicate an improvement for the datasets \textsc{CrisisLex} and \textsc{HumAID} which share similar information type ontologies.
However, the knowledge transfer for \textsc{TrecIS} does not maintain improvements or even leads to worse results.
We believe the reason for this observation is two-fold.
Firstly, the results of \tablename~\ref{table:results:overall} suggest that the obtained embedding representations are less semantically discriminative than the pre-trained language model for multi-label classification.
This may result into a worse cross-corpus knowledge transfer.
Secondly, the information type ontologies of \textsc{CrisisLex} and \textsc{HumAID} differ from \textsc{TrecIS}.
While \textsc{CrisisLex} and \textsc{HumAID} share most of the information types, the ontology of \textsc{TrecIS} differs with fine-grained information types such as \textsc{New Sub Event}, \textsc{EmergingThreats}, and \textsc{Factoid}.

\paragraph{Cross-Lingual}

In the cross-lingual setup, we aim to analyse the adaptation to the German language.
However, we are not aware of any German datasets which cover crisis events and information types.
Therefore, we adopt the \textsc{German BASF Explosion} \cite{habdank2017} and \textsc{German Floods} \cite{reuter2015} datasets which represent binary classification tasks about relevancy.
We train a classifier on the entire \textsc{CrisisLex} corpus and map the information type prediction \textsc{NOT RELATED} to the irrelevant class and all other categories to the relevant class.
Here, we assume that the SE considers irrelevant and relevant clusters in the embedding space which can boost the relevancy classification.
Furthermore, we compare the multilingual variant of \textsc{MPNET\textsubscript{SE}}\footnote{\url{sentence-transformers
/paraphrase-multilingual-mpnet-base-v2}} and the translation\footnote{\url{Helsinki-NLP/opus-mt-en-de}} to English tweets.
We summarize the findings in \tablename~\ref{table:results:crosslingual} whereby \textsc{Random} corresponds to a randomized classifier.
As expected, the comparison of the \textsc{Random} baseline and the MLP classifiers validates the task transfer to the binary classification.
Importantly, we experience the effectiveness of CTS in the cross-task transfer by comparing the language model and CTS fine-tuned SE's.
The performance improvements with the multilingual variant of \textsc{MPNET\textsubscript{SE}+CTS} further demonstrates the capabilities of fine-tuning in the multi-lingual setting.
However, the translated posts outperform the multilingual model in all English model variants.
This is an indication of catastrophic forgetting which occurs during training with only English task data.
The lack of multi-lingual data in the domain of information type classification is still a challenge.

\begin{table}[t]
    \centering
    \resizebox{\linewidth}{!}{
    \begin{tabular}{lccc}
        \toprule
        \multicolumn{1}{c}{} & \multicolumn{3}{c}{\textbf{Target Corpus}} \\
        \cmidrule(rl){2-4} 
        & \textbf{\textsc{CrisisLex}} & \textbf{\textsc{HumAID}} & \textbf{\textsc{TrecIS}} \\
        \toprule
        \textbf{\textsc{CrisisLex}} & - & 63.7\hspace{0.3mm}$_{3.8}$ (3.6$\uparrow$) & 22.9\hspace{0.3mm}$_{2.5}$ (0.0$\phantom\uparrow$) \\
        \textbf{\textsc{HumAID}} & 50.4\hspace{0.3mm}$_{5.9}$ (1.4$\uparrow$) & - & 21.9\hspace{0.3mm}$_{2.2}$ (1.0$\downarrow$) \\
        \textbf{\textsc{TrecIS}} & 45.8\hspace{0.3mm}$_{4.1}$ (3.2$\downarrow$) & 56.5\hspace{0.3mm}$_{2.0}$ (0.4$\uparrow$) & - \\
        \bottomrule
    \end{tabular}}
    \caption{High-data results of cross-corpus transfer with \textsc{MPNET\textsubscript{SE}+CTS}. For the results, we report the event macro-averaged $F_1$ (x100\%) score with standard deviations. The numbers and arrows in brackets indicate absolute improvements ($\uparrow$) or degradations ($\downarrow$) in comparison to \textsc{MPNET\textsubscript{SE}}.}
    \label{table:results:crosscorpus}
\end{table}

\begin{table}[t]
    \centering
    \begin{tabular}{lcc}
        \toprule
        \textbf{Language} & \textbf{\textsc{de}} & \textbf{\textsc{de} $\rightarrow$ \textsc{en}} \\
        \toprule
        \textsc{Random} & 42.2 & - \\
        \cmidrule(rl){2-3}
        \textsc{MPNET\textsubscript{LM}} & 48.8 & \underline{54.6} \\
        \textsc{MPNET\textsubscript{SE}+CTS} & 50.1 & \textbf{55.1} \\
        \textsc{MPNET\textsubscript{SE}+ML+CTS} & 53.4 & 54.1 \\
        \bottomrule
    \end{tabular}
    \caption{Results of cross-lingual transfer with \textsc{MPNET} variants trained on \textsc{CrisisLex}. For the results, we report the event macro-averaged $F_1$ (x100\%) score. The column \textsc{de} $\rightarrow$ \textsc{en} indicates the translation to the English language. The symbol \textsc{+ML} represents the multilingual variant of \textsc{MPNET} which is further trained with \textsc{CTS}.}
    \label{table:results:crosslingual}
\end{table}

\section{Conclusion}
\label{sec:conclusion}

In this work, we investigated the contrastive task-specialization of SE's for the information type classification.
The transformation of universal SE's into task-specialized models demonstrates performance gains especially in low-resource setups.
Furthermore, we demonstrate the first results and opportunities in cross-corpus, cross-lingual, and cross-task transfer in the focused crisis domain.
There are multiple avenues for future research that can improve different aspects with respect to information type classification.
Research directions may include but are not limited to: \textbf{1)} data augmentation techniques, \textbf{2)} retrieval-augmented classification, and \textbf{3)} hierarchical contrastive learning.

\section*{Ethical Considerations}

Open Source Intelligence (OSINT) has become a significant role for various authorities and NGOs for advancing struggles in global health, human rights, and crisis management \citep{bernard_2018,evangelista_2021,kaufhold_2021}.
Following the view of OSINT as a tool, our work pursues the goal to support relief government agencies, organizations, and other stakeholders during ongoing and evolving disaster events.
We argue that Natural Language Processing techniques for OSINT and disaster response can have a positive impact on comprehensive situational awareness and in decision-making processes such as coordination of particular services. 
For example, NLP for social media can enrich the information with the public as co-producers \citep{li_2018}.
In contrast, relying on noisy user-generated content as an information source runs the risk of introducing mis- and disinformation.
This can cause adverse effects on downstream processing and requires strategies and particular care before the deployment.
Furthermore, data privacy issues may arise due to the inherited properties of user-based data.
Various anonymization processes should be taken into account for identifying and neutralizing sensitive references \citep{medlock_2006}.

\section*{Limitations}

We believe there is much room for improving the contrastive task-specialization method with respect to the multi-class, multi-label, and noisy user-generated text setup.
We tested only one variant of language models without considering different transformer encoder sizes or specialized Twitter-based pre-trained models.
Furthermore, we did not conduct experiments for a variety of loss functions which may be expanded to triplet loss, cosine similarity, supervised contrastive loss, and the hierarchical variants for higher level information types.
We relied for the cross-lingual analysis only on the German language datasets and only conducted experiments for relevancy classification which poses a simplification of information type classification.
Future research should consider multi-lingual information type datasets and tasks to comprehensively validate the cross-lingual and cross-task setups in the crisis-related domain.
As highlighted in section \ref{sec:conclusion}, sophisticated data augmentation methods can further improve the overall classification results but still poses a major challenge for noisy user-generated content.
Lastly, recent advanced in the area of instruction-based Large Language Model's (LLM's) should be considered for future research.

\section*{Acknowledgements}

The authors acknowledge the financial support by the Federal Ministry of Education and Research of Germany in the project ISAKI (project number 13N15572).

\bibliography{anthology,custom}
\bibliographystyle{acl_natbib}

\end{document}